\documentclass{article}

\usepackage{graphicx}
\usepackage{times}
\usepackage{wrapfig}
\usepackage{color}
\usepackage{t1enc}
\usepackage{subfigure}
\usepackage{amsfonts}
\usepackage{graphicx}
\usepackage{hyperref} 
\usepackage{amsmath} 

\usepackage{graphicx}
\usepackage{graphics}
\usepackage{enumitem}
\usepackage[below]{placeins}
\usepackage{arxiv}

\usepackage[utf8]{inputenc} 
\usepackage[T1]{fontenc}    
\usepackage{hyperref}       
\usepackage{url}            
\usepackage{booktabs}       
\usepackage{amsfonts}       
\usepackage{nicefrac}       
\usepackage{microtype}      
\usepackage{lipsum}

\title{TWO TIER PREDICTION OF STROKE\\ USING ARTIFICIAL NEURAL NETWORKS AND SUPPORT VECTOR MACHINES}

\author{
  Jerrin Thomas Panachakel\thanks{Based on the Master's project done by the Jerrin Thomas Panachakel under the guidance of Jeena R.S. Full report available at \url{https://sites.google.com/view/jerrinpanachakel/publications}} \\
  Dept. of Electrical Engineering\\
  Indian institute of Science, Bangalore\\
  India \\
  \texttt{jerrinp@iisc.ac.in} \\
   \And
 Jeena R.S. \\
  Dept. of Electronics and Communication Engineering\\
  College of Engineering, Trivandrum\\
  India\\
  \texttt{jeenars@cet.ac.in} \\
}

\begin{document}
\maketitle

\begin{abstract}
Cerebrovascular accident \textit{(CVA)} or stroke is the rapid loss of brain function due to disturbance in the blood supply to the brain. Statistically, stroke is the second leading cause of death. This has motivated us to suggest a two-tier system for predicting stroke; the first tier makes use of Artificial Neural Network (ANN) to predict the chances of a person suffering from stroke. The ANN is trained the using the values of various risk factors of stroke of several patients who had stroke. Once a person is classified as having a high risk of stroke, s/he undergoes another the tier-2 classification test where his/her neuro MRI (Magnetic resonance imaging) is analysed to predict the chances of stroke. The tier-2 uses Non-negative Matrix Factorization and Haralick Textural features for feature extraction and SVM classifier for classification. We have obtained an accuracy of 96.67\% in tier-1 and an accuracy of 70\% in tier-2.
\end{abstract}


\section{Introduction}
Cerebrovascular accident (CVA) or cerebrovascular  insult  (CVI), commonly referred to as ``stroke'' is the leading cause of death, next to  ischaemic heart disease and the leading cause of adult disability worldwide \cite{f1,f2}. Globally, 15 million people suffer from stroke every year and of this, a third dies and half of the remaining struggle with permanent disabilities \cite{f5}. The statistics is no different for developing countries like India \cite{f6,f9,f8}. In Trivandrum, the capital of the Indian state of Kerala, the incidence rate of stroke is 135.0 in the urban community and 138.0 in the rural community \cite{f9}. Clearly, stroke has transformed from a disease pertaining to developed nations to a global hazard. 

The steps taken by major health organisations such as 1) WHO, 2) AHA (American Heart Association), 3) Indian Stroke Association etc. for reducing the stroke fatalities and permanent disabilities can be broadly classified into two:
\begin{enumerate}
\item Improvements in the stroke treatment methods, development and optimisation of early diagnosis techniques and training of personnels to handle stroke cases effectively.
\item Establishing stroke risk factors through scientific research and analysis which can be used for preventing the incidence of stroke by creating awarenesses in the public and by administering treatments for reducing the risk of having a stroke. 
\label{as}
\end{enumerate}

The former step is not entirely applicable to low-income and middle-income countries due to the constraint in the resources, both in terms of trained personnels and money. But the results of the latter step can be put to work, irrespective of the availability of money. But this approach still requires trained personnels who is aware of the stroke risk factors and how much each for the risk factor can contribute to the total risk, both individually and collectively. For instance, in India, which is a lower-middle-income country according to The World Bank \cite{f11}, there is only one neurologist for every 3 million people \cite{f12,f13,f14}. Not only that, these neurologists are heavily laden with the treatment of non-stroke disorders \cite{f12}.

In this work, we propose a two tier system for the prevention of stroke. The first tier makes use of stroke risk factors, much the same way mentioned earlier, except for the fact that instead of trained professionals, machine learning to used. An ANN (Artificial Neural Network) is trained using the stroke risk parameter values of subjects who had stroke and risk parameter values of normal subjects. This can be viewed as a regression problem \cite{f15,f16} with the output of the system giving a score that indicates the net stroke risk of the patient.

In the tier-2 of the proposed system, we make use of neuroimaging, feature extraction and classification techniques to give an additional information on the risk of the person to have a stroke. A multilevel Support Vector Machine (SVM) trained on T2-weighted MRI images of subjects who had stroke and the same of normal subjects is used to classify a given T2-weighted MRI image into two classes:either having a high risk of stroke or a low risj of stroke. Similar systems for predicting neurological disorders such as Alzheimer's disease can be found in the literature \cite{40,a4}. In our work, we used Haralick features \cite{a6} and  Non-negative Matrix Factorisation (NMF) \cite{48} for feature extraction. To the best of our knowledge, this is a maiden work in the field of stroke prediction. We also propose a novel multilevel classification system which can incorporate two non-linear features in such a way that their combination gives better classification efficiency than what can be achieved if there are used individually or if used by simple concatenation.

\section{Prior Art}
\begin{figure}
\centering
\includegraphics[width=5in]{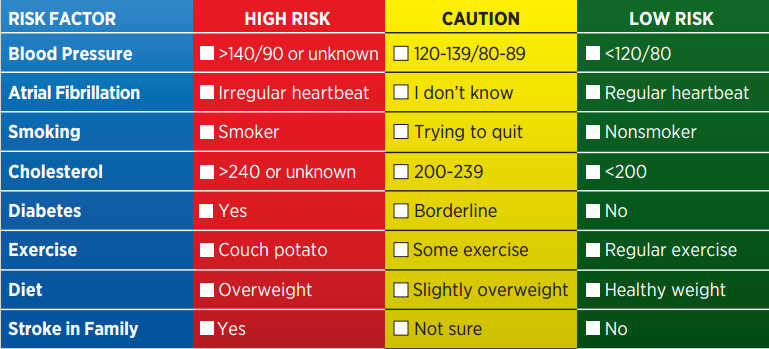}
\caption{Stroke risk factors.}
\label{7}
\end{figure}

As such, there are not many works in the literature where stroke is predicted using neuroimaging. Nevertheless, there are some works, for instance the one by National Stroke Association, USA, where they have developed a score card as shown in Fig.\ref{7}. A score of $3$ or more on this scorecard means that the person has a high probability of getting suffering from a brain attack. 

Lloyd \textit{et al.} in \cite{41} have proposed a method for predicting IS (Ischemic Strokes) by considering non-traditional factors such as body mass index, waist:hip ratio, high density lipoprotein cholesterol, albumin, von Willebrand factor, alcohol consumption, peripheral arterial disease, and carotid artery wall thickness. But they could achieve only a modest improvement in thee prediction capability compared to models that use only traditional risk factors. 
\subsection{Stroke risk factors}

The stroke Association of UK classifies the stroke  risk factors into two groups \cite{n10}:
\begin{enumerate}
\item Lifestyle risk factors
\item Medical conditions risk factors
\end{enumerate}
\subsubsection{Lifestyle risk factors}
The following are included in lifestyle risk factors:
\begin{enumerate}
\item Smoking habits
\item Diet
\item Overweight and obesity 
\item Exercise habits
\item Alcohol consumption
\end{enumerate} 
\subsubsubsection{Smoking habits}
Smoking has a significant effect on stroke risk factor of a subject. According to \cite{82,83,n10}, smoking doubles the risk of stroke. A person smoking 20 cigarettes a day has six times the risk of stroke compared to a non-smoker \cite{87}. For a person with high blood pressure, the risk to have a stroke is 5 times more than a smoker with normal blood pressure and 20 times more than a non-smoker with normal blood pressure \cite{88}. \cite{89} reports that 10\% of deaths from stroke are due to smoking.
\subsubsubsection{Diet}
High amounts of fruits and vegetables in diet can reduce the risk of stroke by up to 30\%. Higher the amount of salt intake, higher will be the risk of stroke \cite{94}. There will be a 23\% increase in the risk for an increase of 5g of salt consumption a day \cite{94}.
\subsubsubsection{Others}
An overweight person has a 22\% higher risk of ischemic stroke whereas an obese person has a 64\% higher risk \cite{98}. When moderate physical activity reduces the risk of stroke by upto 27\%, physical inactivity can increase the risk by 50\% \cite{101,102}. Regular alcohol consumption can can result in a three fold increase in the risk of stroke \cite{105}.
\subsubsection{Medical Conditions}
Various medical condition risk factors include:
\begin{enumerate}
\item Atrial fibrillation
\item High blood pressure
\item Diabetes
\item High cholesterol
\end{enumerate}
\subsubsubsection{Atrial fibrillation}
Atrial fibrillation (AF) is the irregular or abnormal heart rhythm. AF can increase the risk of stroke by a factor of five \cite{116}. Treatment of AF can prevent stroke to a great extent since a a fourth of the people who have a stroke have AF too.
\subsubsubsection{Others}
High blood pressure is a leading cause of stroke. A half of the stroke are caused due to high blood pressure \cite{107}. A person who is diabetic has twice the risk of having a stroke when compared to a normal person \cite{112}. High cholesterol in conjunction with smoking habits or physical inactivity has a significant effect on risk of a person to have stroke. Reducing cholesterol can reduce the risk of stroke by 21\% \cite{121}.

\cite{11} gives a slightly different set of stroke risk factors which are:
\begin{enumerate}
\item Hypertension
\item Diabetes mellitus
\item Smoking habits
\item Dyslipidemia
\item Atrial fibrillation
\end{enumerate}
The authors have also given a result which supports this work; more than 94\% of the subjects who had stroke has at least one of these stroke factors above the normal value. 
\section{Dimensionality reduction techniques}
\subsection{Motivation}
Quite often, we are often faced with the task of handling very high dimensional data in various applications\footnote{It may be noted that the term ``high'' is not objective but a subjective term. It all depends on several factors like hardware specifications, application, required response time etc. For instance, a few  Megabytes of data can be ``high'' for a system that has only a FDD (floppy disk drive) as the storage system (eg.Yamaha PSR 1100- an arranger workstation of the previous decade) but will be a ``low'' for a system that has a HDD (hard disk drive) .}. This problem can be posed mathematically as follows: The given obervation vector which is ``high'' dimensional in nature is denoted as {$\boldsymbol{x}$}. The vector can be viewed as a member of $p$ dimensional vector space and hence can be written as 
\begin{equation}
\boldsymbol{x}=(x_1,x_2,\ldots ,x_p)^T 
\label{eq1}
\end{equation}.
Now, our task is to create a $k$ dimensional vector $\boldsymbol{s}$, expressed mathematically as:
\begin{equation}
\boldsymbol{s}=(s_1,s_2,\ldots ,s_k)^T 
\label{eq2}
\end{equation}, subject to the following conditions:
\begin{enumerate}[leftmargin=1.2cm]
\item $k\le p$, i.e.,the dimension of $\boldsymbol{s}$ is less than the dimension of $\boldsymbol{s}$
\item $\boldsymbol{s}$, in some criterion, represents $\boldsymbol{x}$
\end{enumerate}  This high dimension of data may make the task in our hand too complex due to several reasons: 
\begin{enumerate}
\item Hardware limitations, that include memory imitations, processor power limitations etc. to handle the high dimension. These limitations can also manifest as a power limitation especially for mobile devices since more memory and processor usage means more power consumption.
\item There is a finite probability that as the number of features increases, the classification error initially goes down but at some point, it starts to grow \cite{f20}, which is definitely bad and needs to be avoided.  
\end{enumerate}

The reason why dimensionality reduction is possible is because of the following \cite{f30,f31}:
\begin{enumerate}
\item All practical measurements are corrupted by measurement noise. In a given data, the high variations might be due to the measurement noise and hence many of the variable might be irrelevant. 
\item Some of the variables in the observation vector $\boldsymbol{x}$ given in (\ref{eq1}) can be expressed as a linear combination of other variables in the same vector. In other words, the variables are correlated to each other and the task is to find a new vector $\boldsymbol{s}$, given in (\ref{eq2}) in which the variables are hopefully, uncorrelated. The variables in the new vector are often referred to as ``latent variables'' or ``hidden variables'' \cite{f30,f32,f33,f34}.
\end{enumerate}

\subsection{Non-negative Matrix Factorisation (NMF)}
One of the major drawbacks of PCA is that negative values of the basis vectors are quite difficult for being interpreted in many practical applications. One of remedies for this was to have a representation where non-negativity is imposed by some means. Given a non-negative matrix $\boldsymbol{A}$, the task is to decompose the matrix into two non-negative matrices $\boldsymbol{V}$ and $\boldsymbol{H}$ subjected to the condition that the Forbenius norm between the given matrix $\boldsymbol{A}$ and the product of the two vectors $\boldsymbol{V}\boldsymbol{H}$ is minimum. 
\begin{equation}
\min_{\boldsymbol{V}\ge0,\boldsymbol{H}\ge0}||\boldsymbol{A}-\boldsymbol{V}\boldsymbol{H}||^2_F
\label{12}
\end{equation}
This decomposition is known as  Non-Negative Matrix Factorization (NMF) \footnote{A comprehensive review of NMF can be found in \cite{f35}}. The matrix $\boldsymbol{V}$ is called the basis matrix or the mixing matrix and $\boldsymbol{H}$ represents unknown or hidden sources. The beauty of (\ref{12}) lies in the perspective of viewing each column of $\boldsymbol{A}$ as a linear combination of columns of $\boldsymbol{V}$ with the weights given by the components of each column in $\boldsymbol{H}$ \cite{nn1,nn2,nn3}. The number of columns in $\boldsymbol{V}$ is very much less than the number of rows in $\boldsymbol{A}$. Though NMF seems to be a computationally complex operation, several algorithms have been developed in the last decade for the computationally efficient implementation of NMF.

The maiden work in the field of NMF (Non-negative Matrix Factorisation) can be traced back to a 1994 paper by Paatero and Tapper \cite{17} in which they performed factor analysis
on environmental data \cite{f35}. Their aim was to find the common latent features or latent variables that explained the given set of observation vectors. Some elementary variables combine together positively to give each of the factor.  A factor can either be present, in which case it has a positive effect or the factor can be absent, in which case the factor has null influence. Clearly, there is no room for a ``negative'' influence and hence  it often makes sense to
constrain the factors to be non-negative.

Given a non-negative matrix $\boldsymbol{A}$ whose each columns correspond to different variables of  an observation matrix, the task is to decompose $\boldsymbol{A}$ into another two $\boldsymbol{V}$ and $\boldsymbol{H}$, subjected to the condition that both $\boldsymbol{V}$ and $\boldsymbol{H}$ are also non-negative in nature. The columns of $\boldsymbol{V}$ can be considered to be the factors and the rows of $\boldsymbol{H}$ are the influences of these factors. $\boldsymbol{W}$is used to denote the weight associated to each element, which indicates the level of
confidence in that measurement. Paatero and Tapper advocate optimizing the function
\begin{equation}
||W.(A-VH)||_F^2   \text{  subject to   }  V\ge0\ and\  H\ge0
\end{equation}
\begin{figure}
\centering
\includegraphics[width=5in]{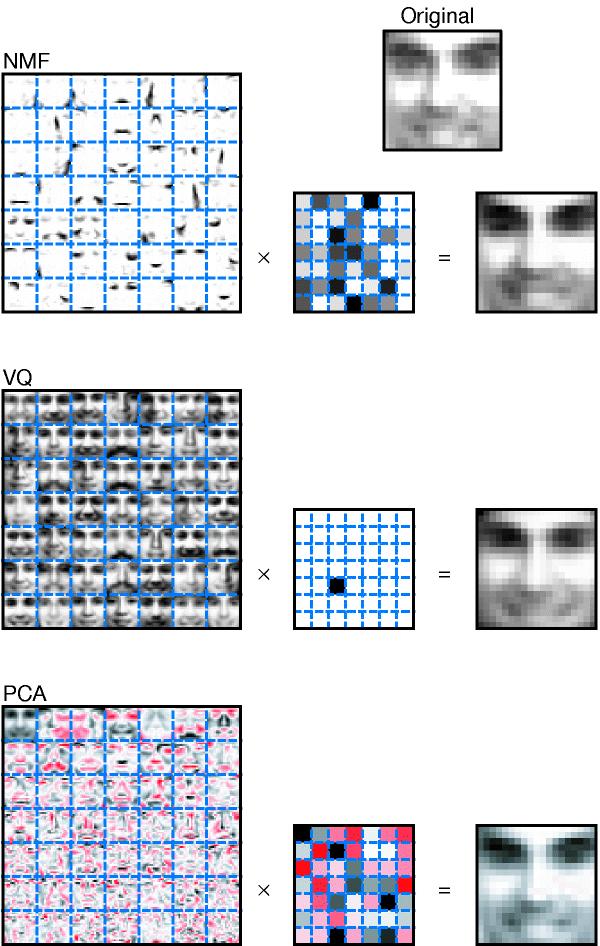}
\caption[Comparison of representation of face using NMF, VQ and PCA .]{Comparison of representation of face using NMF (top), VQ (middle) and PCA (bottom). From \cite{nn6} }
\label{eg}
\end{figure}

Independent of Paatero, Lee and Seung introduced
the concept of NMF in a 1997 paper titled\  ``Algorithms for Non-negative Matrix Factorisation''\ \cite{19,f35}. They begin by considering
the following encoding problem. Suppose that the columns of $\boldsymbol{V}$ are fixed feature vectors
and that $\boldsymbol{A}$ is an input vector to be encoded. The task is to minimize the reconstruction error $||\boldsymbol{A}-\boldsymbol{V}\boldsymbol{h}||_2^2$. Mathematically, 
\begin{equation}
\min_{\boldsymbol{h}}||\boldsymbol{A}-\boldsymbol{V}\boldsymbol{h}||_2^2
\end{equation}
Depending on the constraint on $\boldsymbol{H}$, different learning algorithms can be developed. The choice for unconstrained minimisation is PCA. Contrary to PCA, Vector Quantization (VQ) requires that
$\boldsymbol{h}$ equal one of the canonical basis vectors (i.e. a single unit component with the remaining
entries zero). Lee and Seung proposed a convex coding scheme which requires the entries of $\boldsymbol{h}$ to be non-negative numbers which
sum to one.  So the encoded vector is the best approximation to the input from the convex
hull of the feature vectors. This was one of the two techniques which they put forward. The second technique which they put forward was conic coding. This  requires  that the entries of $\boldsymbol{h}$ be
non-negative, like in the case of the convex coding scheme but doesn't have the constraint that the sum of the elements in $\boldsymbol{h}$, which are non-negative, to be unity. Then the encoded vector is the best approximation to the input from the cone
generated by the feature vectors.

\section{Textural features}
Haralick Textural features were developed by R.M. Haralick in the year 1973 \cite{a6}. The basis for these features is the gray-level co-occurrence matrix, $\mathbf{G}$ given by,
\begin{equation}
\mathbf{G}=\left[
\begin{array}{cccc}
p(1,1) & p(1,2) & \cdots & p(1,N_g) \\
p(2,1) & p(2,2) & \cdots & p(2,N_g) \\
\vdots & \vdots & \ddots & \vdots   \\
p(N_g,1) & p(N_g,2) & \cdots & p(N_g,N_g) \\
\end{array}
\right]
\end{equation} 
 This is a square matrix with dimension $N_g$, where $N_g$ is the number of gray levels in the image. Element $[i,j]$ of the matrix is generated by counting the number of times a pixel with value $i$ is adjacent to a pixel with value $j$ and then dividing the entire matrix by the total number of such comparisons made. Each entry is therefore considered to be the probability that a pixel with value $i$ will be found adjacent to a pixel of value $j$.

 \begin{figure}
 \centering
 \includegraphics[width=2.7in]{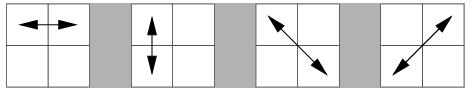}
 \caption[Four directions of adjacency.]{ Four directions of adjacency as defined for calculation of the Haralick texture features. The Haralick statistics are calculated for co-occurrence matrices generated using each of these directions of adjacency.}
 \label{2}
 \end{figure}
 Since adjacency can be defined to occur in each of four directions in a 2D, square pixel image (horizontal, vertical, left and right diagonals - see Fig. \ref{2}), four such matrices can be calculated.
 
 Haralick then described 14 statistics that can be calculated from the co-occurrence matrix with the intent of describing the texture of the image: 
 \begin{enumerate}
 \item Angular Second Moment:
 \begin{equation}
  \sum_i\sum_jp(i,j)^2
 \end{equation}
 \item Contrast: \begin{equation}
 \sum_{n=0}^{N_g-1}n^2\{\sum_{i=1}^{N_g}\sum_{j=1}^{N_g}p(i,j)\},|i-j|=n
 \end{equation}
 \item Correlation: \begin{equation}
 \frac{\sum_i\sum_j(ij)p(i,j)-\mu_x\mu_y}{\sigma_x\sigma_y}
 \end{equation}
 \item Sum of Squares: Variation: \begin{equation}
 \sum_i\sum_j(i-\mu)^2p(i,j)
 \end{equation}
 \item Inverse Difference Moment: \begin{equation}
 \sum_i\sum_j\frac{1}{1+(i-j)^2}p(i,j)
 \end{equation}
 \item Sum Average: \begin{equation}
 \sum_{i=2}^{2N_g}ip_{x+y}(i)
 \end{equation}
 \item Sum Variance: \begin{equation}
 \sum_{i=2}^2N_g(i-f_s)^2p_{x+y}(i)
 \end{equation}
 \item Sum Entropy:\begin{equation}
  -\sum_{i=2}^{N_g} p_{x+y}(i)\log\{p_{x+y}(i)\}
 \end{equation}
 \item Entropy:\begin{equation}
  -\sum_i\sum_jp(i,j)\log (p(i,j))
 \end{equation}
 \item Difference Variance:\begin{equation}
  \sum_{i=0}^{N_g-1}i^2p_{x-y}(i)
 \end{equation}
 \item Difference Entropy: \begin{equation}
 -\sum_{i=0}^{N_g-1}p_{x-y}(i)\log\{p_{x-y}(i)\}
 \end{equation}
 \item Information Measure of Correlation 1: \begin{equation}
 \frac{HXY-HXY1}{\max\{HX,HY\}}
 \end{equation}
 \item Information Measure of Correlation 2: \begin{equation}
 (1-\exp[-2(HXY2-HXY)])^{0.5}
 \end{equation}
 \item Maximal Correlation Coefficient: \begin{equation}
 \text{second largest eigen value of Q}^{0.5}
 \end{equation}
 \end{enumerate}
 \qquad where $\mu_x$,$\mu_y$, $\sigma_x$ and $\sigma_y$ are the mean and standard deviations of $p_x$ and $p_y$, the probability density functions. ${Q}(i,j)$ is given by the following relation:  
 \begin{equation}
 {Q}(i,j)=\sum_k\frac{p(i,j)p(j,k)}{p_x(i)p_y(k)}.
 \end{equation}

\section{}{Database}
Two datasets were used in this work: one which contains neuroimages of 30 subjects and the other which contains the values of stroke risk parameters of 30 subjects. These datasets are explained below,
\section{Neuroimages dataset}
This dataset work mainly includes the images from ``The Whole Brain Atlas''\footnote {The database can be obtained from {\url{www.med.harvard.edu/aanlib/home.html}.}} \cite{121}, developed by Keith Johnson, MD, and Alex Becker, PhD., with the support of  the Brigham and Women's Hospital Departments of Radiology and Neurology, Harvard Medical School, the Countway Library of Medicine, and the American Academy of Neurology. The database includes the MRI images of neurological disease such as:
\begin{enumerate}
\item Neoplastic Disease (brain tumor)
\begin{itemize}
\item Metastatic adenocarcinoma
\item Metastatic bronchogenic carcinoma
\item Meningioma
\item Sarcoma
\end{itemize}
\item Degenerative Disease
\begin{itemize}
\item Alzheimer's disease
\item Huntington's disease
\item Motor neuron disease
\item Cerebral calcinosis
\end{itemize}
\item Inflammatory or Infectious Disease
\begin{itemize}
\item Multiple sclerosis
\item AIDS dementia
\item Creutzfeld-Jakob disease
\item Cerebral Toxoplasmosis
\end{itemize}
\item Cardiovascular Accident (CVA)
\begin{itemize}
\item Acute stroke: Speech arrest 
\item Acute stroke: ``alexia without agraphia'' 
\item Subacute stroke: ``transcortical aphasia'' 
\item Chronic subdural hematoma
\item Hypertensive encephalopathy, and
\item Cerebral hemorrhage.
\end{itemize}
\end{enumerate}
\begin{figure}
\centering
\subfigure[Database]{
\includegraphics[width=5in]{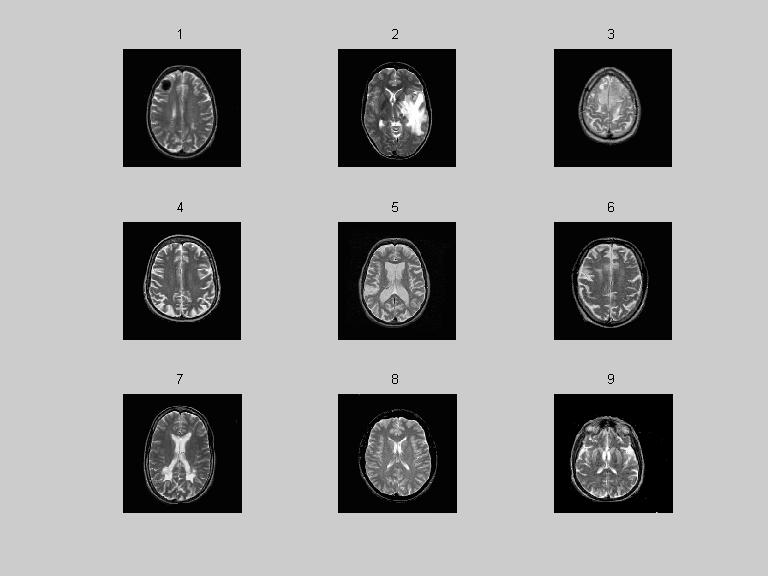}}
\centering
\subfigure[Database cont.]{
\includegraphics[width=5in]{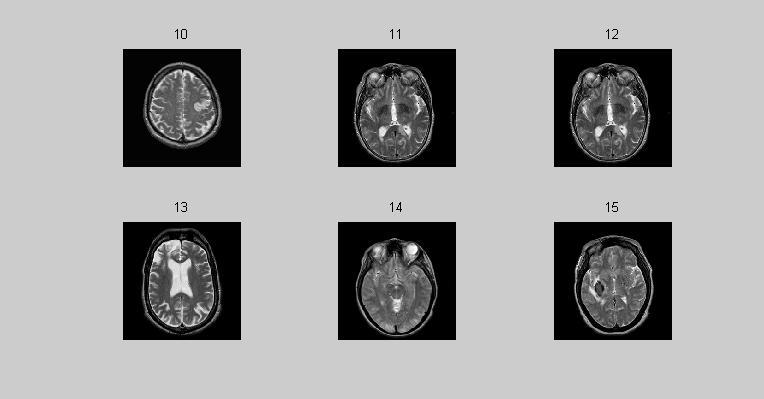}}
\caption{Database for neurological disorder classifier}
\label{5}
\end{figure}
T2-weighted MRI images of 30 subjects were used for this work. In these 30 images, 14 were of subjects who had stroke and 16 were normal subjects. A part of the database  is given in Fig. \ref{5}. The images $10$ to $15$ corresponds to stroke images and they are 10: Acute stroke: Speech arrest, 
11: Acute stroke: ``alexia without agraphia'', 
12: Subacute stroke: ``transcortical aphasia'', 
13: Chronic subdural hematoma,
14: Hypertensive encephalopathy, and
15: Cerebral hemorrhage.

\begin{figure}
\centering
\subfigure[Original MRI]{
\includegraphics[width=4in]{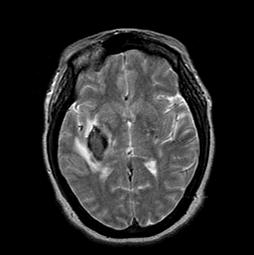}}
\centering
\subfigure[Preprocessed MRI]{
\includegraphics[width=4in]{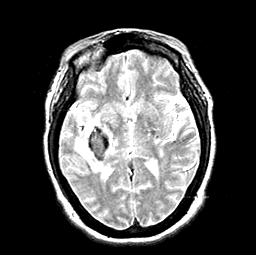}}
\caption{Original and preprocessed images}
\label{50}
\end{figure}
Prior to feature extraction, the images were subjected to both anatomic and intensity normalisation. Normalising the images with respect to the maximum pixel intensity value may introduce noise \cite{40,lbs1}, so the maximum pixel intensity was considered to be the mean of the highest $0.1 \%$ pixel intensity values, as proposed in \cite{40} and \cite{a7}.
Original and preprocessed images are shown in Fig. \ref{50}.
This database will be hereafter referred to as ``TBA'' database.
\section{Risk parameter database}
This database was obtained from the data obtained from a private medical college in Trivandrum, kerala. It contains the stroke risk parameter values of 30 subjects, 15 of whom had stroke and the rest have a very low stroke risk score according to the stroke risk card of American Stroke Association. Stroke risk factors which were considered as:
\begin{enumerate}
\item Blood pressure- in mmHg
\item Atrial fibrillation- yes/no
\item Smoker- yes/no
\item Cholesterol levels- mg/dL
\item Diabetic- yes/no
\item Exercise habits- yes/no
\item Obese- yes/no
\item Stroke in family- yes/no
\end{enumerate}
This database will be hereafter referred to as the ``SRP'' database.

\section{Classification of structural MRIs}
\label{cl}
\subsection{Motivation}
As a predecessor to the proposed CAD (Computer Aided Diagnosis) tool for prediction, we first developed a CAD tool for differentiating brain MRIs obtained from the TBA database of subjects who have suffered from stroke (these MRIs will be hereafter referred to as ``stroke MRI'') from the MRI of subjects suffering from other neural disorders (these MRIs will be hereafter referred to as ``non-stroke MRI''), which can be diagnosed from structural MRIs \cite{panachakel2017multi}. For this, we made use of the Haralick features \cite{a6} and Non-negative Matrix Factorisation \cite{17} for feature extraction and SVM for classification. Also, we have introduced a computationally efficient method for combining feature vectors which are linear in two different kernel spaces. For this, we have made use of the distance of the feature vector from the hyperplane as a measure of confidence value of classification, thus improving the classification efficiency which could be obtained by using either one of the two feature vectors or by mere concatenation of the two features.

The features used were NMF and Haralick features. For NMF, the number of basis was set to 14. The Haralick features used  were the range and mean of these 14 statistical parameters descried in \cite{a6} along the 4 directions as the first set of features, with unity distance between the neighbours. 8 BPP images were scaled down to 4 BPP images to reduce the computation time. 
\subsection{Multi-level SVM}
In the proposed multi-level classification, two support vector models are created using NMF features and Haralick features. Now, given a feature vector $\Phi(\overrightarrow x)$, in the feature space $\Phi(.)$, we compute a score based on its distance from the decision boundary hyperplane $f(x)$ as,\cite{n3}
 \begin{equation}
 \text dist (\Phi(x),f(x))=\frac{f(x)}{\sum\limits_{i \in SV}|y_i\alpha_i\Phi(x)|^2}
 \end{equation} 
 where $y_i\in \{-1,1\}$, $\alpha_i : $ vector weights for support vectors.
 
 The scores provide an estimate of how good the classification is. Larger the score, larger will be the distance from the hyperplane and hence higher will be probability of the sample to lie in that class \cite{n3}. In our work, for a given test sample, we compute the scores for both the support vector models. The model which gives the highest absolute value for the score is assumed to have classified the sample correctly. This approach improves the classification accuracy.
  \subsection{Performance Metrics}
   \begin{figure}[h!]
    \centering
    \includegraphics[width=5in]{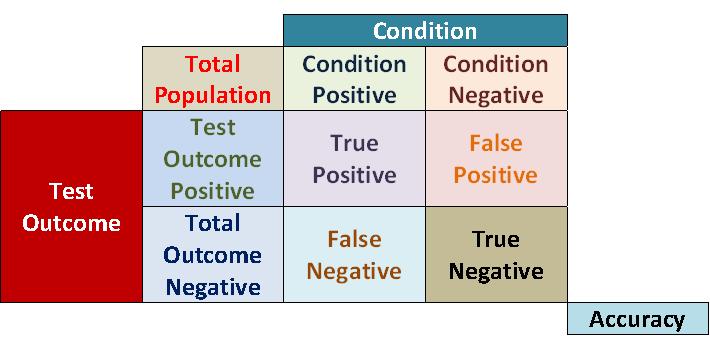}
    \caption{Sample confusion matrix}
    \label{cf1}
    \end{figure}
  \label{s3}
  Three performance metrics were used in this work:
  \begin{enumerate}
  \item Sensitivity ($SN$), which is the measure of the system's ability to identify stroke MRIs.
  \item Specificity ($SP$), which is a measure of the system's ability to identify non-stroke MRIs.
  \item Accuracy ($AC$), which is a measure of the system's net classification efficiency.
  \end{enumerate}
  Before defining these metrics mathematically, we introduce the following terms:
  \begin{itemize}
  \item $TP$: True Positive, stroke MRI identified as stroke MRI.
  \item $TN$: True Negative, non-stroke MRI identified as non-stroke MRI.
  \item $FP$: False Positive, non-stroke MRI identified as stroke MRI.
  \item $FN$: False Negative, stoke MRI identified as non-stroke MRI.
  \end{itemize}
  Now, we can mathematically define sensitivity, specificity and accuracy as:
  \begin{equation}
  SN=\frac{TP}{TP+FN}
  \end{equation}
  \begin{equation}
  SP=\frac{TN}{TN+FP}
  \end{equation}
  \begin{equation}
  AC=\frac{TP+TN}{TN+TP+FP+FN}
  \end{equation}

  For the purpose of cross-validation, LOOCV (Leave-One-Out Cross Validation) was used. As the name suggests, Leave-One-Out Cross Validation (LOOCV) involves using an observation as the validation set and the remaining observations as the training set. This is repeated on all ways to cut the original sample on a validation set of an observation and a training set. 
  
   A sample confusion matrix which will be used for providing the results is shown in Fig. \ref{cf1}.
  \section{Results}
   
    \begin{figure}
    \centering
    \subfigure[Using NMF]{
    \includegraphics[width=5in]{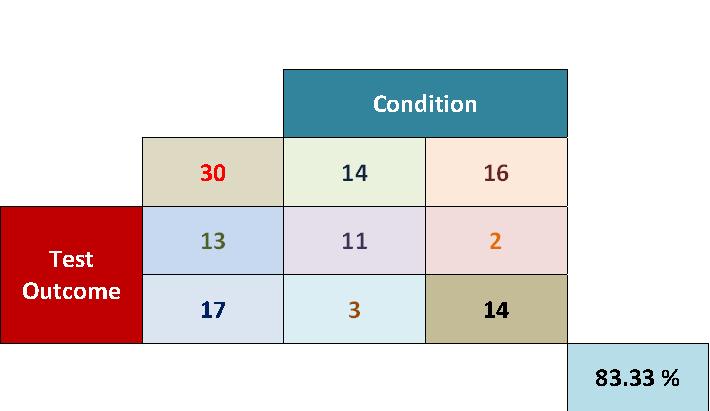}}
    \centering
    \subfigure[Using Haralick features]{
    \includegraphics[width=5in]{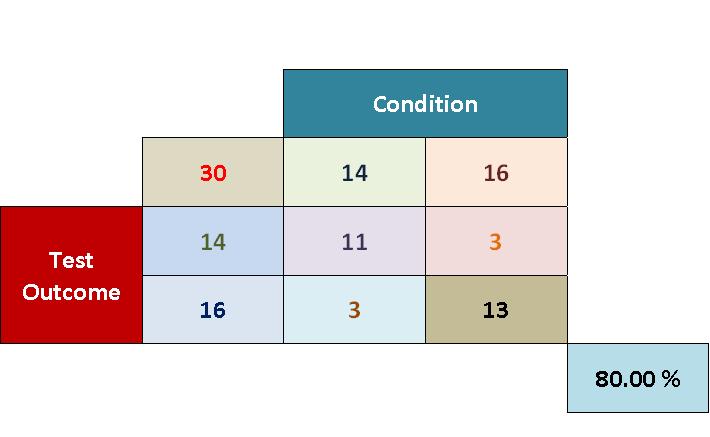}}
    \caption{Confusion matrices for various features considered individually.}
    \label{c5}
    \end{figure}
        \begin{figure}
        \centering
        \subfigure[Simple concatenation]{
        \includegraphics[width=5in]{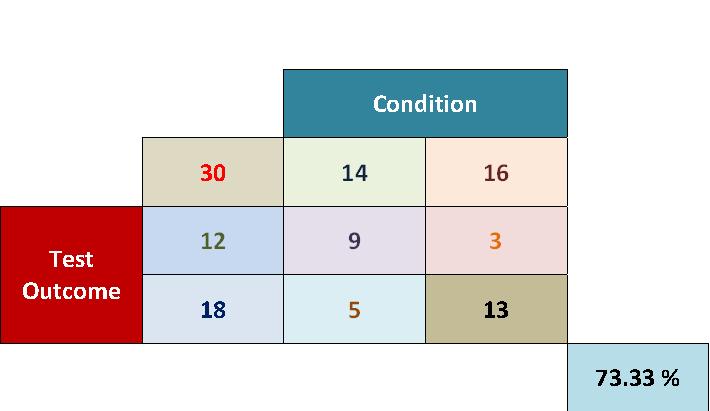}
        \label{w1}}
        \centering
        \subfigure[Multi-level SVM]{
        \includegraphics[width=5in]{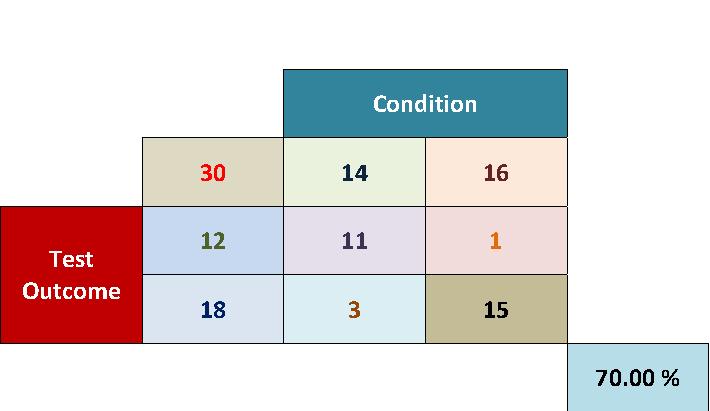}
        \label{w2}}
        \caption{Confusion matrices for various features considered simultaneously.}
        \label{c6}
        \end{figure}
          \begin{figure}[h!]
                \centering
                \includegraphics[width=5in]{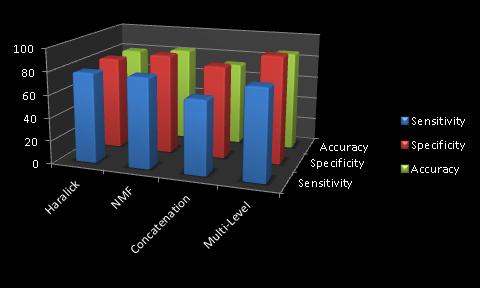}
                \caption{Comparison of multi-level SVM with simple SVM}
                \label{cfe}
                \end{figure}
  \label{res}
  Feature extraction and classification was done in MATLAB\textsuperscript{\textregistered} 2013a. SVM functions in  the statistical toolbox was used in this work. Different kernel functions with different parameter values were tried out in the classification process to obtain the highest possible classification accuracy. 
  
   \begin{table}
   \centering
   \begin{tabular}{|c||c|c|c|}
   \hline  & Linear & MLP & RBF \\ 
   \hline \hline Sensitivity & 71.43 & 78.57 & 71.00 \\ 
   \hline Specificity & 87.50 & 81.25 & 81.25 \\ 
   \hline Accuracy & 80.00 & 80.00 & 76.67 \\ 
   \hline 
   \end{tabular} 
     \caption{Using Haralick features}
          \label{q1}
    \end{table}
  For classification using Haralick features, for RBF (Radial Basis Function) kernel, highest accuracy was obtained for a ``rbf\_sigma'' of 60. For MLP (Multilayer Perception), the parameter was [10000 -100]. The results are given in TABLE \ref{q1}.
  While using NMF, the ``rbf\_sigma'' value for highest efficiency was 40 and the LMP parameter was [10 -100]. The results are shown in TABLE \ref{q2}
     \begin{table}
     \centering
     \begin{tabular}{|c||c|c|c|}
     \hline  & Linear & MLP & RBF \\ 
     \hline \hline Sensitivity & 78.57 & 64.29 & 71.43 \\ 
     \hline Specificity & 87.50 & 87.50& 81.25 \\ 
     \hline Accuracy & 83.33 & 76.67 & 76.67 \\ 
     \hline 
     \end{tabular} 
       \caption{Using NMF}
            \label{q2}
      \end{table}
 
  The corresponding confusion matrices for the best classification output \footnote{When the classification accuracies are equal, the next parameter compared was sensitivity.}is shown in Fig. \ref{c5}. 
  \begin{table}
       \centering
       \begin{tabular}{|c||c|c|c|c|}
       \hline  & Haralick & NMF & Concatenated & Multi-Level \\ 
       \hline \hline Sensitivity & 78.57 & 78.57 & 64.29 & \textbf{78.57} \\ 
       \hline Specificity & 81.25 & 87.50& 81.25 & \textbf{93.75} \\ 
       \hline Accuracy & 80.00 & 83.33 & 73.33 & \textbf{86.67}\\ 
       \hline 
       \end{tabular} 
         \caption{Comparison of multi-level SVM with simple SVM}
              \label{q3}
        \end{table}
  
    Fig. \ref{c6} shows the confusion matrices when the features are considered simultaneously. Fig. \ref{w1} shows the confusion matrix when the two features are simply concatenated. RBF kernel with ``rbf\_sigma'' of 40 was used since it gave the best performance. Fig. \ref{w2} shows the confusion matrix when the proposed multi-level SVM is used.
     
        Fig. \ref{cfe} shows a graphical comparison of sensitivity, specificity and accuracy when the features are used individually and simultaneously.

Clearly, multi-level SVM outperform ordinary SVM. The developed system has a sensitivity of 78.57\% at a specificity of 93.75\% yielding an accuracy of 86.67\% as given in TABLE \ref{q3}. The analysis showed that false detection occurred mainly for  neoplastic disorders.

\section{Prediction of CVA}
\subsection{Tier 1}
       \begin{figure}[h!]
        \centering
        \includegraphics[width=4in]{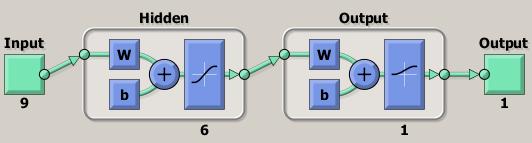}
        \caption{Neural Network (NN) used in Tier-1}
        \label{nn}
        \end{figure}

                    \begin{figure}[h!]
                        \centering
                        \includegraphics[width=5in]{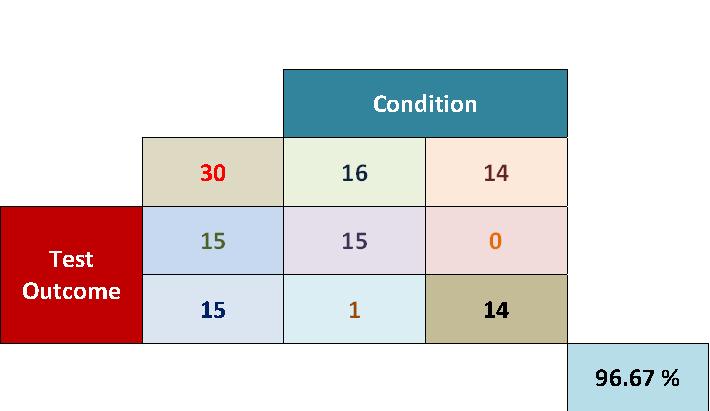}
                        \caption{Confusion matrix of Tier-1}
                        \label{cnn}
                        \end{figure}

Tier 1 makes use of the SRP database. For classification, it uses FFNN (Feed-Froward Neural Network), with nine input nodes, one hidden layer with six neurons and two output neurons. The training algorithm used is Levenberg-Marquardt backpropagation algorithm and the learning rate was 0.1 with the rate to decrease the learning rate set to 0.5. These parameters were found to be the most appropriate after intense training and cross validation tests with several parameter combination. This neural network is shown in Fig.\ref{nn}.

                        The  confusion matrix obtained for tier-1 is shown in Fig. \ref{cnn}.
\subsection{Tier 2}
The tier-2 makes use of neuroimaging and machine learning for predicting CVA. The ideal source of dataset for testing this tier would have been the brain MRI images of subjects who had stroke, taken before the onset of stroke. But, it is very difficult to obtain such a dataset. So instead, the brain MRI images of patients who had stroke were taken from the TBA database and the all lesions due to stroke, visible in them were removed with the help of a group consisting of a general physician and a radiologist. The portions removed were assigned ``NaN'' (Not a Number) as the pixel intensity values and the MATLAB\textsuperscript{\textregistered} routine for evaluating the Haralick features was modified so that pixels with intensity values of NaN were completely neglected in the formation of the covariance matrix. 28 features as described in Section \ref{res} were used for the classification.   

During the training stage, normal brain MRIs from the TBA database were used but for the testing and cross validation stages, the brain MRIs with lesions removed were used. Due to the removal of  lesions, NMF could not be used, and hence this stage relies entirely on Haralick features.       

As with the ``Classification of structural MRIs'', described in Section \ref{cl}, different SVM with different kernel functions were used to ``predict'' stroke. This is called ``prediction'' because once the dead brain cells are removed, the MRI does not have any lesions of a stroke and the MRI, as such, corresponds to a normal MRI. Since the MRI is a ``stroke MRI'' (ref. Section Section \ref{cl}), it is reasonable to assume that the MRI corresponds to a subject who is going to have a stroke, i.e., the stage before there are any typical stroke symptoms visible in the MRI. It is a reasonable because stroke is usually a focal brain ischemia and results in the death of brain cells ina particular area where the blood flow is blocked \cite{122}. If the proposed classifier for predicting stroke can classify a ``stroke MRI'' with the lesions removed, it means that tthere are are some latent features in the brain MRI of a stroke patient other than the stroke lesions which the features used have successfully extracted. 

    \begin{table}
     \centering
     \begin{tabular}{|c||c|c|c|}
     \hline  & Linear & MLP & RBF \\ 
     \hline \hline Sensitivity & 100 & 85.71 & 100.00 \\ 
     \hline Specificity & 06.25 & 56.25& 06.25 \\ 
     \hline Accuracy & 50.00 & 70.00 & 50.00 \\ 
     \hline 
     \end{tabular} 
       \caption{Using NMF}
            \label{q22}
      \end{table}

                          \begin{figure}[h!]
                              \centering
                              \includegraphics[width=5in]{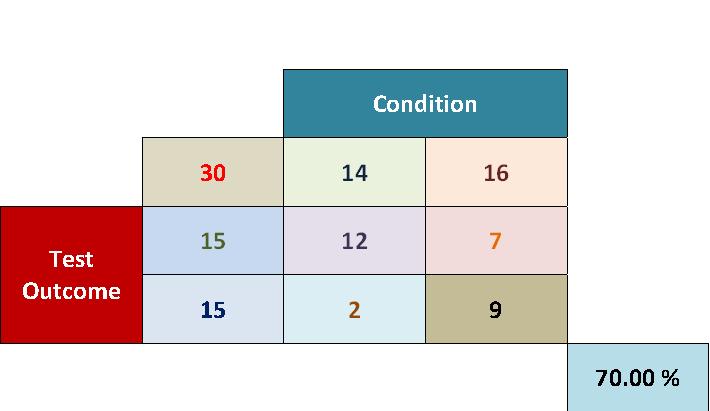}
                              \caption{Confusion matrix of Tier-2}
                              \label{t2}
                              \end{figure}
The classifier performance for various kernel functions is given in TABLE \ref{q22}. The various parameters were: 
\begin{itemize}
\item RBF: 20
\item MLP: [1 -2.54]
\end{itemize}
The confusion matrix when the kernel function was MLP is given in FIg. \ref{t2}.

Even though. a classification efficiency of 70\% may not be very high, considering the fact that only Haralick features were used and the sensitivity is greater than  85\% at a specificity greater than 55\%. This can definitely be improved by incorporating other features too. In that case, this can be considered as a promising results. 
\section{Conclusion}
This work introduced two CAD systems, one for classifying structural MRI images and other for predicting CVA. A novel technique for classifying using separate set of features which are linear in different kernels were also introduced. This technique, named ``multi-level SVM'' improves the classification accuracy by 18 \% over a classifier that uses the two features simultaneously by concatenating them together.  

The accuracy for classifying structural MRI images using NMF, Haralick features and multi-level SVM was 86.67\%. This acted  as the prelude to a CAD tool for predicting CVAs. The CAD tool for prediction is two tier in architecture. The first tier makes use of ANN and stroke risk factor parameters for creating a system, which can predict the probability of a person to have a stroke, given his/her stroke risk parameter values.

The second tier makes use of neuroimaging for the prediction. T2-weighted structural MRI images were used for training the system. The term ``prediction'' is used because the input to the system during the testing stage has the stroke lesions removed. The classification accuracy obtained by using Haralick features alone was 70\%. This shows that there are changes in the brain other than the lesions caused by oncosis. The classification efficiency can be improved by using other features or by improving the classifier. These CAD tools can be used for predicting the stroke risk of a person, thus reducing the incidence rate of CVA,which is the second leading cause of death.  

\bibliographystyle{IEEEtran}
\bibliography{IEEEabrv,thes}
\end{document}